\documentclass[10pt,twocolumn,letterpaper]{article}

\usepackage{iccv}
\usepackage{times}
\usepackage{epsfig}
\usepackage{graphicx}
\usepackage{amsmath}
\usepackage{amssymb}

\usepackage{bbm}
\usepackage{booktabs} 
\usepackage{cuted}
\usepackage{capt-of}
\usepackage{subfig}
\usepackage[font=small,skip=0pt]{caption}
\usepackage[breaklinks=true,bookmarks=false]{hyperref}

\iccvfinalcopy % *** Uncomment this line for the final submission
\def\iccvPaperID{****} % *** Enter the ICCV Paper ID here

\ificcvfinal\fi

\begin{document}

%%%%%%%%% TITLE

\title{\vspace{-3em}Two-Stage Monte Carlo Denoising with Adaptive Sampling and Kernel Pool}

\author{

Tiange Xiang\thanks{This work is done when Xiang was an intern at Tencent AI Lab.}\\
The University of Sydney\\
{\tt\small txia7609@uni.sydney.ed.au}
\and
Hongliang Yuan\\
Tencent AI Lab\\
{\tt\small haroldyuan@tencent.com}

\and
Haozhi Huang\\
Tencent AI Lab\\
{\tt\small huanghz08@gmail.com}

\and

Yujin Shi\\
Tencent AI Lab\\
{\tt\small yujinshi@tencent.com}
\vspace{-3em}
}

% \author{

% Tiange Xiang\thanks{This work is done when Tiange Xiang was an intern at Tencent AI Lab.}, Hongliang Yuan, Haozhi Huang and Yujin Shi\\
% The University of Sydney\\
% Tencent AI Lab\\
% {\tt\small txia7609@uni.sydney.edu.au}\\
% {\tt\small \{haroldyuan, mathzhuang, yujinshi\}@tencent.com}
% \vspace{-3em}
% }

\maketitle

\begin{strip} \centering
\includegraphics[width=1.0\linewidth]{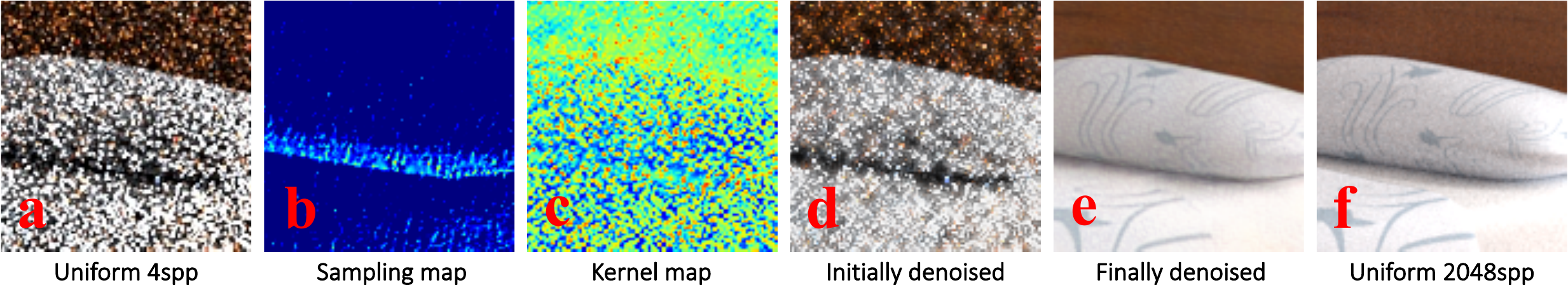}
\captionof{figure}{Instead of denoising on Monte Carlo path traced images with uniform samples (a) directly, our framework functions on adaptively sampled ones guided by a neural network inferred sampling map (b). A  kernel map (c) is concurrently predicted to promote an initial denoising on the adaptively sampled images (d) before feeding into a novel second stage denoiser. Compared to  reference image rendered in high spp (f), our final denoised result (e) recovers most texture and highlighting details.}

% \captionof{figure}{We propose an adaptive sampling framework that performs two-stage denoising on Monte Carlo path traced images.}
\label{fig:fig1}
\end{strip}

\ificcvfinal\thispagestyle{empty}\fi

%%%%%%%%% ABSTRACT
\begin{abstract}
Monte Carlo path tracer renders noisy image sequences at low sampling counts. Although great progress has been made on denoising such sequences, existing methods still suffer from spatial and temporary artifacts. In this paper, we tackle the problems in Monte Carlo rendering by proposing a two-stage denoiser based on the adaptive sampling strategy. In the first stage, concurrent to adjusting samples per pixel (spp) on-the-fly, we reuse the computations to generate extra denoising kernels applying on the adaptively rendered image. Rather than a direct prediction of pixel-wise kernels, we save the overhead complexity by interpolating such kernels from a public kernel pool, which can be dynamically updated to fit input signals. In the second stage, we design the position-aware pooling and semantic alignment operators to improve spatial-temporal stability. Our method was first benchmarked on 10 synthesized scenes rendered from the Mitsuba renderer and then validated on 3 additional scenes rendered from our self-built RTX based renderer. Our method outperforms state-of-the-art counterparts in terms of both numerical error and visual quality.

\end{abstract}

%%%%%%%%% BODY TEXT
\section{Introduction} \label{intro}
Restoring degraded visual signals with noise is of immense importance to the computer vision and graphics community. With the recent success of deep learning algorithms, great progress has been made on denoising images \cite{Quan_2020_CVPR, Xia_2020_CVPR} and videos \cite{Yue_2020_CVPR, Tassano_2020_CVPR, Ehret_2019_CVPR}. As one of the most common degradation types in digital images, injected Gaussian noise \cite{nam2016holistic}, can be regarded as stochastic samples in a latent Gaussian distribution and can be easily captured by well-designed convolutional neural networks. However, capturing videos under low intensity light circumstances usually introduce \textit{realistic} noise with more complex implicit patterns. A naive apply of standard Gaussian noise removal methods on such realistic noise will not suffice \cite{xu2018external, xu2018trilateral, xu2017multi}. 

As a simulation of real-world light transmission, ray tracing introduces a large amount of realistic noise when the light sampling count is limited. Current Monte Carlo path tracers randomly sample lighting directions on hemispheres modeled on the collision of scene objects with the emitted lights, makes the decision of lighting path from the light source to the observer a fully stochastic process. When the number of samples per pixel (spp) is considerably low, scarce lighting information causes a great amount of realistic noise. Therefore, a pertinent denoiser to this specific situation is desired to handle the unstructured noise and smooth the significant degradation in the rendered scenes.

% However, the motion of observers and objects in the scene posts a huge challenge to the temporal stability, and additional pre-processing or post-possessing is therefore inevitable \cite{arias2018video, buades2016patch, liu2010high}

Recent advances on denoising path traced videos depend on recurrent variants \cite{vogels2018denoising, chaitanya2017interactive, kaplanyan2019deepfovea} of U-Net \cite{ronneberger2015u} that are fed with video frames iteratively to capture better temporal information. Despite the careful handling of temporal disparities \cite{arias2018video, buades2016patch, liu2010high}, denoised frames still suffer from loss of reconstruction details, especially at dark regions and object contours. Reconstruction failures are blamed on the insufficient visual clues provided in the raw rendered image where lights are hard to reach or transmit through. Monte Carlo path tracing at interactive sampling rates \cite{kuznetsov2018deep, hasselgren2020neural} serves as a good remedy to such problems. Instead of sampling light rays uniformly, the ray number per pixel is dynamically determined through an additional neural network. Although such adaptive sampling methods perform well on hard-to-denoise pixels, there are two major flaws: (1) As a trade-off to the increasing performance, overall computational costs inevitably doubled due to the introduction of a second network. (2) Distributing a limited sampling budget to all frame pixels undermines the visual quality at low-frequency regions, and consequently harms spatial stability.

In this work, our three-fold contributions are made to overcome the above flaws. \textbf{(1)} We rethink the adaptive sampling framework by reusing the computations in the sampling network to function as an initial denoising process. Apart from the determination of sampling count per pixel, we learn an affiliate set of denoising kernels to achieve an initial denoising. Differing from existing Kernel Prediction (KP) \cite{vogels2018denoising, hasselgren2020neural} methods that spawn one-to-one kernel for each pixel, we minimize the exceeding computation overhead by maintaining a public \textit{kernel pool} with a fixed number of communal kernels. A pixel-wise kernel map is then produced from the sampling network to guide the interpolation of pixel-wise kernels in the kernel pool. \textbf{(2)} We propose an adaptive pooling scheme, namely \textit{position-aware pooling} to smooth spatial instabilities by taking positional information into consideration. Additionally, instead of aligning motions for better temporal stability, we introduce \textit{semantic alignment} that learns feature-wise alignment between the current and precedent frame. \textbf{(3)} Our method was validated on 10 different scenes rendered from the Mitsuba path tracer \cite{Mitsuba} and 3 additional scenes rendered from our self-built RTX-based renderer. Our method surpasses other counterparts both quantitatively and qualitatively.

\section{Related Works}

\subsection{Denoising for Monte Carlo Rendering}

Denoiser for Monte Carlo path tracer enables the renderer to operate in a low sampling count environment without an apparent loss on image quality, saving rendering costs. Here, we focus on reviewing the denoising methods that reconstruct degraded visual signals in the image-space.

% Denoiser design for Monte Carlo path tracer is of immense importance to the graphics community \cite{zwicker2015recent}, enabling the renderer to operate in a low sample count setting without an apparent loss on image quality, therefor saving rendering costs. Existing denoising methods reconstruct degraded visual signals via either pixel-wise or sample-wise approaches.

% \noindent
% \textbf{Pixel-based methods.} 
Image-space algorithms manipulate discordant color units and reconstruct noisy pixels directly. Starting from utilizing a set of pre-defined filters \cite{kalantari2013removing, rousselle2012adaptive} to work as a weighted linear combination of the noisy pixels at local windows, similar approaches have been experimented with bilateral filters \cite{rousselle2013robust, sen2012filtering} and non-local patches \cite{moon2013robust}. Instead of combining pixels directly, efforts have been made on generating pixel residuals based on local regressors \cite{bitterli2016nonlinearly, moon2014adaptive}. The above techniques usually require heavy manual analysis of statistical estimations between the denoised image and reference. The image quality improvement brought by such methods is limited and manually tuned parameters on one scene can hardly generalize to others. 

With the thrive of deep learning, neural network based denoisers have been recently more favored. A pioneer practice \cite{kalantari2015machine} utilizes a basic multi-layer network to take image-space features as input and predict the parameters for denoising bilateral filters. To smooth the temporal artifacts in a sequence of noisy rendered images, \cite{chaitanya2017interactive} designed a recurrent neural network that denoises rendered video frames based on the iterative forward of the network. However, the disparity between adjacent frames impacts the performance of the recurrent denoiser greatly, and complex image-space motion alignments are desired. Our proposed method follows a similar recurrent paradigm but handles both spatial and temporal instabilities from novel perspectives.

% \noindent
% \textbf{Sample-based methods.} Differing from direct pixel manipulation, other methods \cite{hachisuka2008multidimensional, lehtinen2011temporal, lehtinen2012reconstructing, gharbi2019sample} generate noise-free images by reconstructing radiance functions with the help of pre-rendered samples. Standing out as a deep learning based method, \cite{gharbi2019sample} designed a dedicated network that takes raw rendering samples as inputs and generates a set of dynamic splatting kernels from samples to color pixels. Unlike gathering filters that aggregate local neighbors, splatting filters distribute center values to nearby units. However, such sample-based methods require larger amount of computations and are empirically inefficient. Also, the designed modules yield poor flexibility and are hard to be plugged into existing rendering pipelines.

\begin{figure*}[t]
	\begin{center}
		\includegraphics[width=1.0\linewidth]{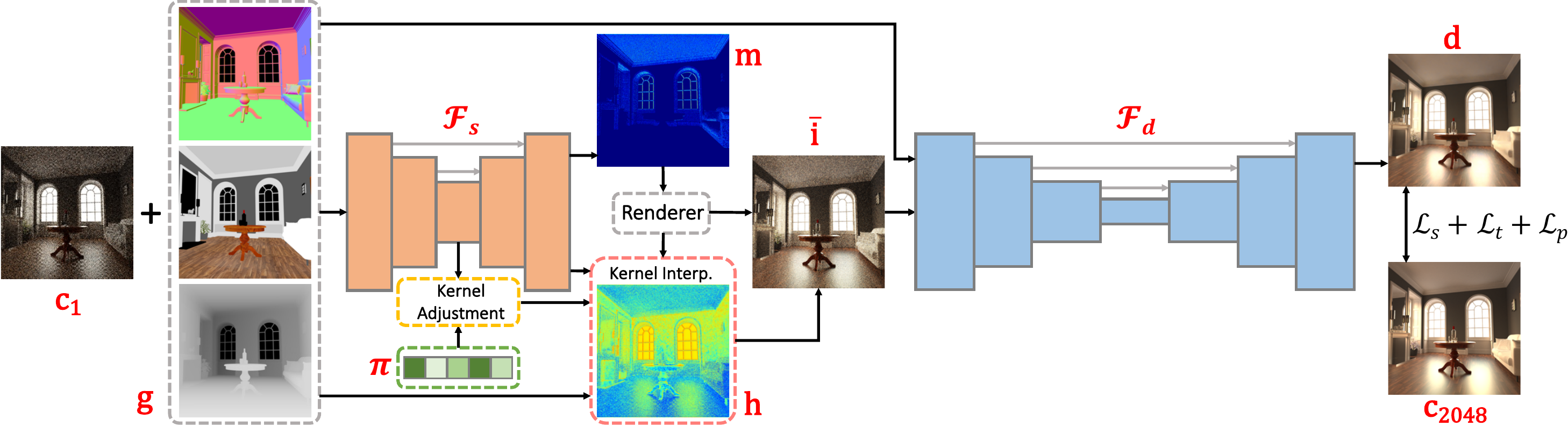}
	\end{center}
	\caption{\textbf{Overview of the proposed framework.} Our framework consists of a sampling network $\mathcal{F}_s$ followed by a denoising network $\mathcal{F}_d$. $\mathcal{F}_s$ determines adaptive spp and performs an initial denoising process on the adaptively rendered image, which is then fed into $\mathcal{F}_d$ along with additional supports to obtain the final denoised image.}
% 	\caption{\textbf{Overview of the proposed framework.} We denote $\mathcal{F}_s$ the sampling network; $\mathcal{F}_d$ the denoising network; $\mathbf{c}_1$ the noisy image rendered with uniform 1 spp; $\mathbf{g}$ the rasterized Gbuffers containing shading normal, albdeo, and scene depth; $\pi$ the proposed kernel pool; $\mathbf{i}$ the initially denoised adaptively sampled noisy image; $\mathbf{c}_{2048}$ the reference rendered with uniform 2048 spp; $\mathbf{d}$ the output denoised image.}
	\label{fig:framework}
\end{figure*}

\subsection{Adaptive sampling methods}
An alternative strategy to alleviate denoising difficulties is to adjust pixel-wise sampling counts dynamically, promoting the renderer to generate more samples on the hard-to-denoise regions and fewer samples to the easier ones.

Early attempts distribute samples adaptively based on the measured errors between the noisy and the reconstructed image pair \cite{hachisuka2008multidimensional, overbeck2009adaptive}. The same strategy has also been adopted in \cite{vogels2018denoising}, however, instead of calculating errors directly, the error map is obtained as the prediction of a neural network. Spp is then iteratively decided in an offline manner. For better adaptation into online rendering, \cite{kuznetsov2018deep} utilizes an additional network to predict sampling map beforehand, based on an initially rendered 1spp image, and then consume the estimated sampling map in the major rendering process. Both networks can be optimized jointly using the final reconstruction loss. In a recent work, \cite{hasselgren2020neural} handles the temporal artifacts in videos by warping the previously reconstructed frame to initiate current frame denoising. The denoiser is further studied to predict hierarchical denoising kernels at multi-scales. However, the above framework highly depends on motion alignments which can be easily influenced by off-screen pixels and object occlusions. In this work, we reuse computations in the sampling network to serve as a two-stage denoising framework, and design Novel modules to efficiently acquire large receptive fields and smooth spatial-temporal instabilities.

\section{Preliminary} \label{preliminary}

 Consider a Monte Carlo path tracer rendered noisy image $\mathbf{c}_{\mathbf{n}}$ with uniform spp of $\mathbf{n}$ and a corresponding set of Geometry Buffers (GBuffers) $\mathbf{g}$ consists of scene depth, shading normal, and albedo at primary hit position. The conventional adaptive sampling framework \cite{hasselgren2020neural, kuznetsov2018deep} takes as input $(\mathbf{c}_1, \mathbf{g})$, and eventually outputs the denoised image $\mathbf{d}$.
 
%  The sampling network $\mathcal{F}_{s}$ tasks as input $(\mathbf{c}_1, \mathbf{g})$ and outputs the sampling map $\mathbf{m}$ that holds integers in a specific range indicating the adaptive sample count for each pixel. The rendering process is then triggered again with the guidance from $\mathbf{m}$ and generates the adaptively sampled noisy image $\mathbf{i}$. Finally, the denoising network $\mathcal{F}_d$ consumes $(\mathbf{i}, \mathbf{g})$ and outputs the 

% $\mathbf{m}$ can be predicted by $\mathcal{F}_{s}(\mathbf{c}_1, \mathbf{g})$ and is expected to hold integers within a pre-set range indicating the adaptive sample count for each pixel. However, the immediate output $\bar{\mathbf{m}}$ of $\mathcal{F}_{s}$ carries only floating point values and may distribute across a wide range.

The framework is comprised of two networks including a sampling network $\mathcal{F}_{s}$ and a denoising network $\mathcal{F}_{d}$, both share the same architecture. The sampling network aims at generating a sampling map $\mathbf{m}$ before the main rendering process initiates. To obtain the valid integer sampling map from the predicted floating values, a simple normalization scheme is followed as suggested in \cite{kuznetsov2018deep}:

\begin{equation} \label{eq1}
    \mathbf{m}=\texttt{round}(\texttt{softmax}(\bar{\mathbf{m}})\cdot \mathbf{N}\cdot \mathbf{s}),
\end{equation}
where $\mathbf{N}$ is the total number of pixels and $\mathbf{s}$ is the expected average spp budget. $\mathbf{m}$ is then fed into the Monte Carlo path tracer acquiring the adaptively sampled image $\mathbf{i}$. However, the involvement of renderer at each iteration of network training is empirically inefficient. To accelerate network training, the rendering process is simulated by pre-rendering a set of images $\mathbf{c}_{2^{\{0,\cdots,t\}}}$ at different uniform spp \cite{kuznetsov2018deep} and assemble the simulated $\mathbf{i}$ via:

\begin{equation}
    \mathbf{i}^p = \frac{\sum^{t}_{i=0}\mathbf{c}_{2^i}^p \cdot 2^{i} \cdot \mathbbm{1}(\mathbf{m}^p, 2^{i})}{\mathbf{m}^p},
\end{equation}
where $\mathbbm{1}(\cdot)$ is an indication function that returns 1 if the binary encoding of $\mathbf{m}^p$ is activated at the $i_{th}$ term. After simulating $\mathbf{i}$ from $\mathbf{m}$, the gradients for $\texttt{softmax}(\bar{\mathbf{m}})$ is calculated with the help of the ground truth image $\mathbf{c}_{\infty}$ rendered at high spp: $\frac{\mathbf{c}_{\infty}-\mathbf{i}}{\mathbf{m}}$. Note that $\mathbf{c}_{\infty}$ here is only used for gradient approximation. Both $\mathbf{i}$ and $\mathbf{g}$ are then combined to be fed into the denoising network $\mathcal{F}_d$ which directly predicts the denoised image $\mathbf{d}$.

% However, since $\mathbf{m}$ cannot be directly supervised and Eq. \ref{eq1} is not differentiable, we 

\section{Methods}

Our proposed framework follows the conventional adaptive sampling paradigm that entails a sampling network $\mathcal{F}_{s}$ followed by a denoising network $\mathcal{F}_d$. Both of the networks are constructed in the U-Net \cite{ronneberger2015u} style, such that encoded embeddings at each level are skipped to the decoder for better feature aggregation and gradient preserving. Differing from \cite{kuznetsov2018deep, hasselgren2020neural} that $\mathcal{F}_d$ and $\mathcal{F}_s$ share the identical network architecture, we put less computation resources on $\mathcal{F}_s$ and focus more on $\mathcal{F}_d$ with deeper encoding depth and larger feature width. Unless explicitly specified, the basic building block is designed as the stack of a 3x3 convolutional layer followed by a ReLU non-linearity layer. An overview of the proposed framework is shown in Figure \ref{fig:framework}.

\subsection{Two-stage denoising with kernel pool}

In the existing adaptive sampling paradigm, $\mathcal{F}_s$ contributes little to the final denoising quality except for the generation of $\mathbf{m}$. However, compared to using $\mathcal{F}_d$ as a one-stage denoiser, introducing an extra network inevitably doubles the overall computational costs and slows down the denoising efficiency. We rethink the responsibility of $\mathcal{F}_s$ and refactor the network to carry out an initial denoising process on $\mathbf{i}$ before being fed into $\mathcal{F}_d$. Since the simulation of $\mathbf{i}$ depends on the prediction of $\mathbf{m}$, $\mathcal{F}_s$, therefore, needs to make an inference beforehand. As a result, promoting $\mathcal{F}_s$ as a naive regresser similar to $\mathcal{F}_d$ is impractical.

Parallel to $\mathbf{m}$, we enable the sampling network to generate a set of denoising kernels $\mathbf{K}$ that is subsequently applied on $\mathbf{i}$, obtaining the initially denoised image $\bar{\mathbf{i}}$. However, the commonly adopted pixel-wise kernel prediction method \cite{vogels2018denoising} poses a huge challenge to our $\mathcal{F}_s$: with the same backbone network, the prediction of pixel-wise kernels strongly couples with the prediction of $\mathbf{m}$, thus easily leads to degradation of identical values across all pixels on either sampling map or predicted kernels. Also, large kernels (21x21 in \cite{vogels2018denoising}) are usually too expensive to be predicted, which puts exceeding computation burdens on $\mathcal{F}_s$. 

% Formally, we reformulate the adaptive sampling paradigm as:

% \begin{equation}
% \begin{split}
%     \mathbf{m}, \mathbf{K} &= \mathcal{F}_s(\mathbf{c}_1, \mathbf{g}),\\
%     \mathbf{d} &= \mathcal{F}_d(\texttt{conv}(\texttt{render}(\mathbf{m}), \mathbf{K}),\mathbf{g}),
% \end{split}
% \end{equation}
% where \texttt{conv}($a,b$) applies convolutions on $a$ with pixel-wise kernels $b$ and \texttt{render}($\cdot$) denotes the rendering process.

To overcome the above challenges, instead of a direct prediction of pixel-wise kernels, we learn a public \textit{kernel pool} $\pi$ consisting of a set of $q$ communal kernels with fixed size $l\times l$. The kernel pool can be jointly optimized along with the networks. During network inference, differing from \cite{vogels2018denoising} that predicts per pixel kernel $\in \mathbbm{R}^{\mathbf{N}\times l\times l}$, our $\mathcal{F}_s$ only outputs an extra \textit{kernel map} $\mathbf{h}\in \mathbbm{R}^{\mathbf{N}\times 1}$ indicating the offset of the most appropriate per pixel kernel in the kernel pool. The initial denoising is achieved through two procedures, namely \textit{kernel adjustment} and \textit{kernel interpolation}. Similar to \cite{Tassano_2020_CVPR}, our initial denoising process requires no particular supervision.

% The kernels are then interpolated from $\pi$ given the predicted offset $\mathbf{h}$.

\noindent
\textbf{Kernel Adjustment.} Although the optimized kernel pool fits the training data well, such kernels may generalize poorly on the testing data due to the disparity between training and testing images. Towards better generalization ability and better correspondence to the target image, we learn an auxiliary set of kernel residuals $\mathbf{r}\in \mathbbm{R}^{q\times l\times l}$ to be applied on $\pi$. Rather than a simple addition of $\pi$ and $\mathbf{r}$, sigmoid gated alpha values $\alpha\in \mathbbm{R}^{q\times 1\times 1}$ are additionally learned to achieve kernel-wise alpha blending. In our experiments, we set $q\ll \mathbf{N}$, hence both the generation of $\mathbf{r}$ and the initial denoising process yield only little overhead computations. 

\begin{figure}[t]
	\begin{center}
		\includegraphics[width=1.0\linewidth]{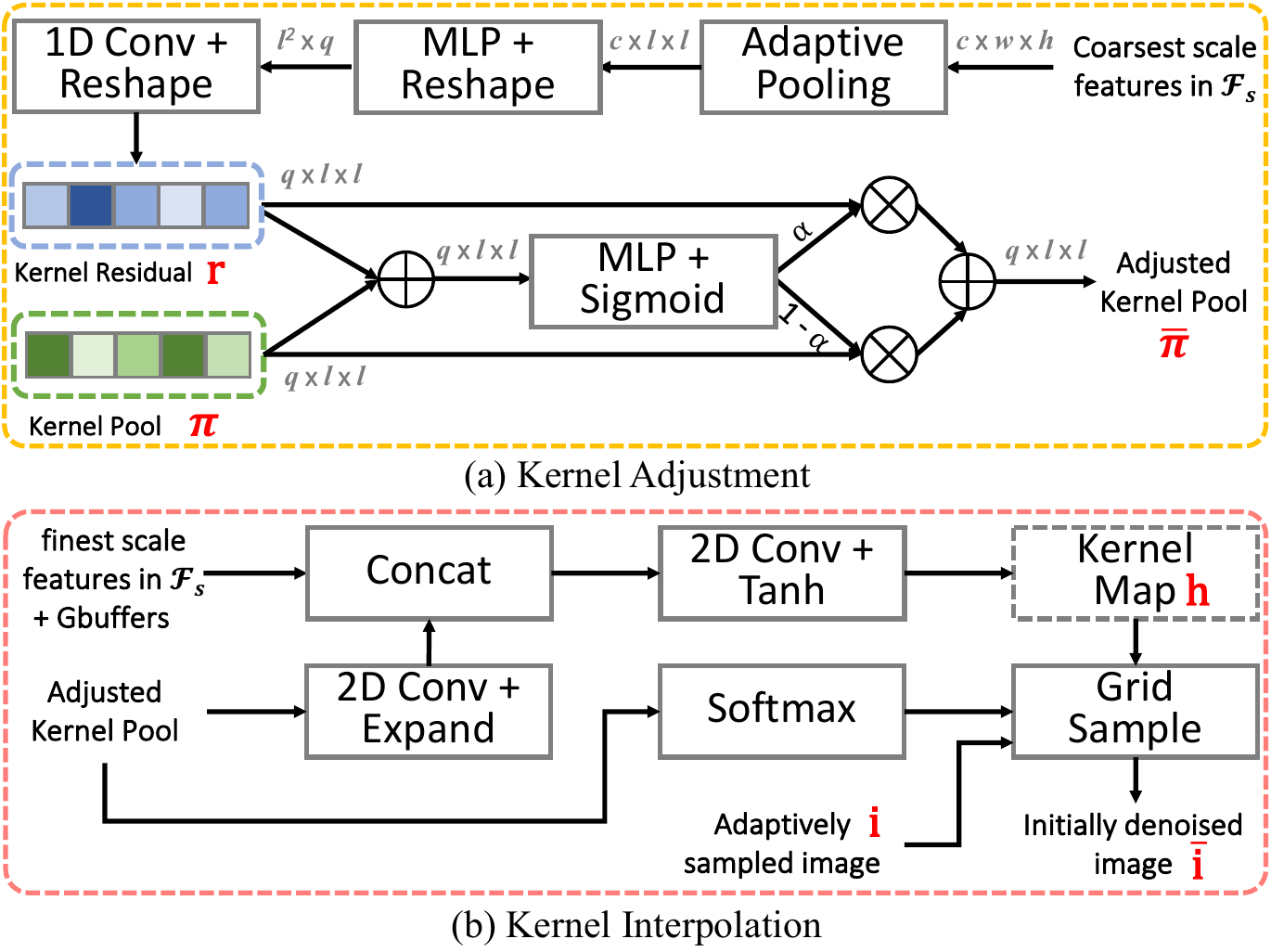}
	\end{center}
	\caption{\textbf{Workflows of the initial denoising procedures.}}
	\label{fig:ka}
\end{figure}

Detailedly, an additional branch is extended from the deepest encoded features in $\mathcal{F}_s$ at the coarsest scale to perform the kernel adjustment separately. The coarsest features are first adaptively pooled to the desired spatial size $l\times l$ which are then excited to $q\times l\times l$ through a MLP. For better correlations between adjacent kernels, we apply an extra convolution with five neighboring kernels at a time sliding from the first kernel to the last obtaining the kernel residual $\mathbf{r}$. We employ an extensive MLP to learn the kernel-wise $\alpha$ scores based on the summation of both $\mathbf{r}$ and $\pi$. The adjusted kernel pool $\bar{\pi}$ for the current input image is then calculated as $\bar{\pi}=(1-\alpha)\pi+\alpha\mathbf{r}$. The kernel adjustment workflow is outlined in Figure \ref{fig:ka}.

\noindent
\textbf{Kernel Interpolation.}  Since $\mathbf{c}_1$ and $\mathbf{i}$ can be different, we predict $\mathbf{h}$ by taking $\mathbf{i}$ as an additional support. Besides, to make $\mathbf{h}$ sensible to $\bar{\pi}$, adjusted kernels are convoluted to a lower dimension with unit spatial size, such kernel descriptor is then spatially expanded and treated as another support. We concatenate the above two supports with the forwarded finest scale features and $\mathbf{g}$ to generate $\mathbf{h}$ through a series of convolutions 
followed by a Tanh layer at the end. 

Per-pixel kernels are grid sampled from $\texttt{softmax}(\bar{\pi})$ with the predicted kernel map $\mathbf{h}$. However, we found that the public kernel pool is not able to satisfy the need for all pixels across an image, and some pixels may rely more on $\mathbf{i}$ rather than $\bar{\mathbf{i}}$ to be further denoised. To this end, we predict an extra offset in $\mathbf{h}$ enabling a 2D interpolation between the softmax gated sampled kernel and an \textit{identity kernel}, which has the same spatial size $l\times l$ but is filled with $1$ at the center and $0$ elsewhere. The updated 2D interpolation enables each pixel in $\bar{\mathbf{i}}$ to be flexibly switched between the initially denoised pixel and the raw adaptively sampled pixel. 

% the prediction of the kernel map relies on $\mathbf{c}_1$ and is not sensitive to the chroma in $\mathbf{i}$, 

\subsection{Spatial-temporal instability smoothing}

\noindent
\textbf{Fast GhostConv.} Towards better distribution of adaptive samples, visual clues implied in $\mathbf{c}_1$ are expected to be well explored. However, path traced samples at low spp pixels may deviate significantly from the true distribution. When determining the light transmission direction via the Monte Carlo method, the stochastically sampled single light ray is likely to have a biased reflection and refraction angle. Consequently, the lighting information rendered in one pixel in $\mathbf{c}_1$ could be represented better in another far-located pixel. To this end, basic operators of $\mathcal{F}_s$ are desired to be designed with larger receptive fields, considering a wide range of pixels at a time thus capturing the deviated ray samples.

We elevate the reception field while further reducing the computational costs by advanced engineering of the Separable Convolution \cite{howard2017mobilenets} and the Ghost Module \cite{han2020ghostnet}. Standard convolutions are first replaced with large-kernel group convolutions to efficiently gather information in a wide range. Different channels in the wide scale features are then communicated through point-wise convolutions. An additional depth-wise convolution is utilized to obtain extra `ghost' features to further widen the receptive field. We denote such efficient operator as \textit{Fast GhostConv} which replace the basic building blocks in $\mathcal{F}_s$ (Figure \ref{fig:operator}).

\begin{figure}[t]
	\begin{center}
		\includegraphics[width=1.0\linewidth]{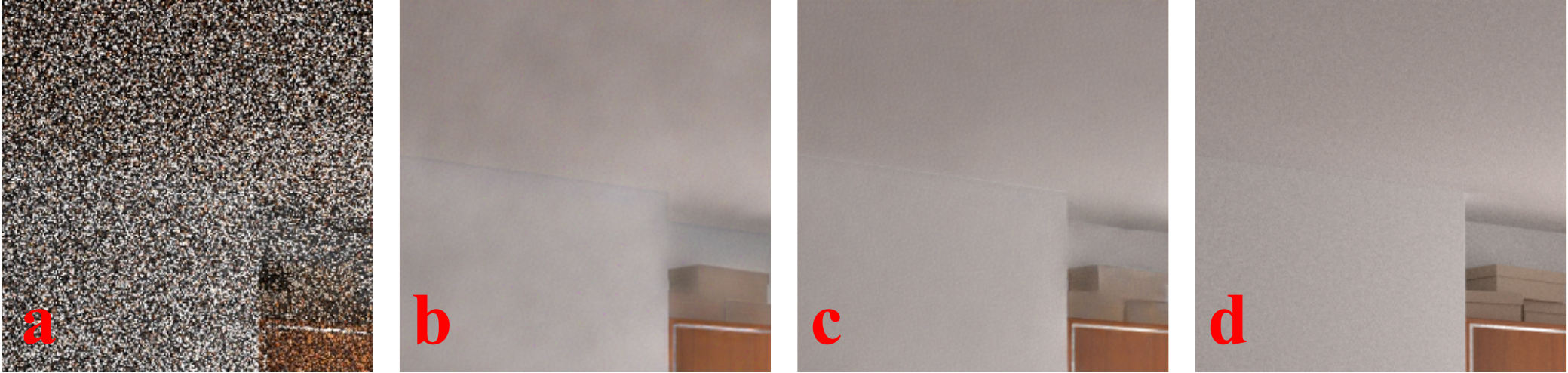}
	\end{center}
	\caption{\textbf{Spatial flicking}. \textbf{(a)} Noisy image $\bar{\mathbf{i}}$. \textbf{(b)} Existing denoising methods may cause severe spatial flicking. \textbf{(c)} Our denoised result. \textbf{(d)} 2048 spp reference. }
	\label{fig:spatial}
\end{figure}

\begin{figure}[t]
	\begin{center}
		\includegraphics[width=1.0\linewidth]{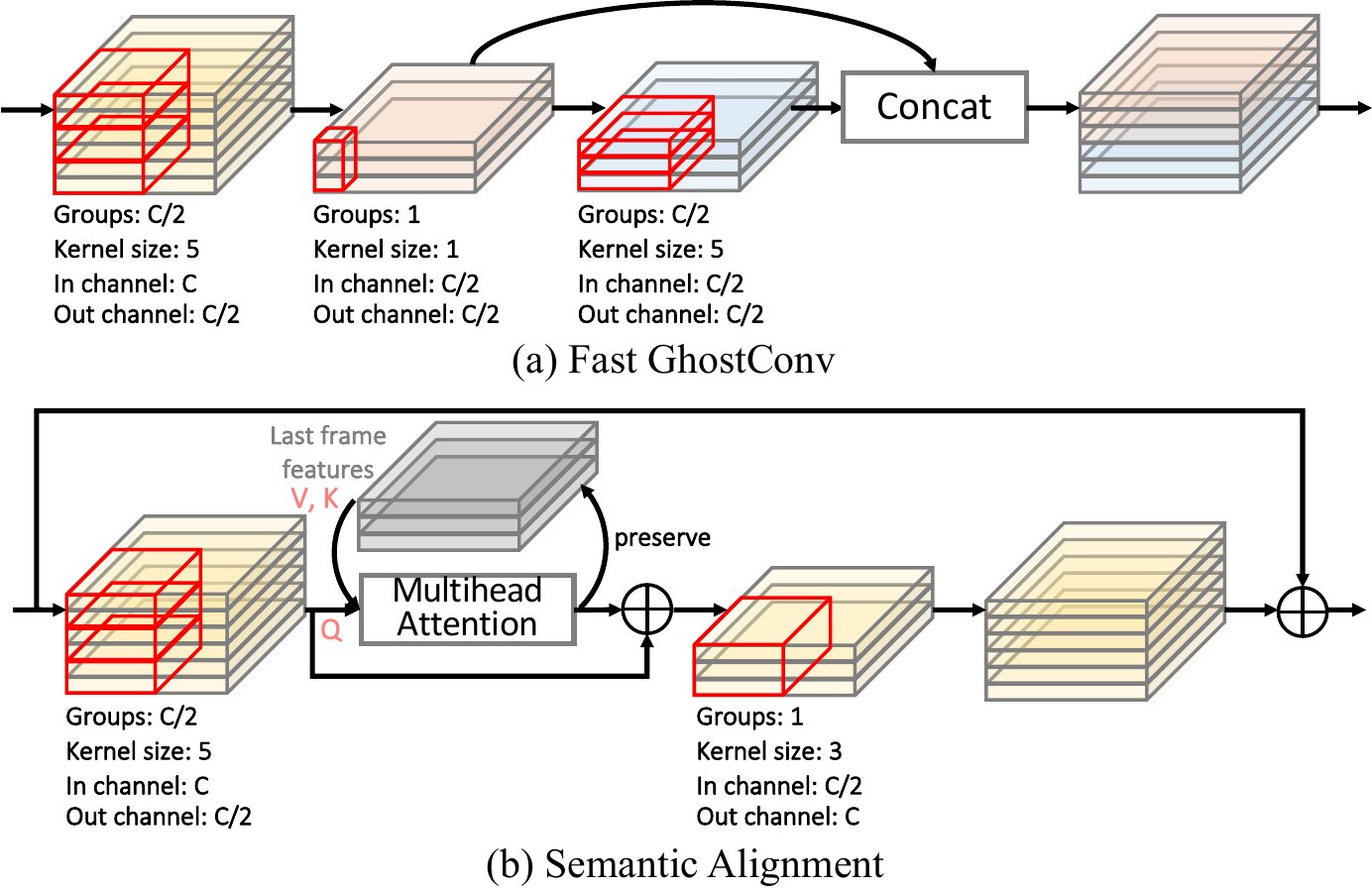}
	\end{center}
	\caption{\textbf{Spatial-temporal instability smoothing}. Convolutions are outlined in red with detailed configures.}
	\label{fig:operator}
\end{figure}

\noindent
\textbf{Position-aware pooling.} Given the fixed total spp budget, assigning more samples to object contours and dark areas results in fewer samples and worse visual quality in low-frequency regions, which occupy most parts of the scenes and could cause conspicuous spatial flickering (Figure \ref{fig:spatial}). We blame such instability on network pooling, where poorly sampled pixels are usually mixed with good ones, and can be hardly discriminate against. 

We learn extra one-to-one scale scores for each member by a cheap MLP in the pooling window to achieve an adaptive pooling. All members in a window are first multiplied with the corresponding scale score and are then summed up altogether to be the pooling result. The scale scores in a window are summed up to 1 and can be gated by a softmax function. Empirically, we found that due to the uncertain directions of ray samples, the proposed pooling strategy is sensitive to positional information. Therefore, we make the adaptive pooling operator position-aware by considering both absolute positions in terms of the whole feature map (encoded as gradient values $\in (-1, 1)$ from the left top corner to the right bottom) and relative positions within the current window (encoded as one-hot vectors).  

\noindent
\textbf{Semantic Alignment.} Inter-frame motions lead to unexpected temporal artifacts. Existing temporal stabilizing methods either estimate pixel-wise motion vectors to align frames directly or consume a sequence of continuous frames at a time. However, the pixel alignment method is sensitive to object occlusions, and consuming consecutive frames demands a large number of extra computations. Considering the above drawbacks, instead of aligning motions in the raw RGB space, we achieve the alignment in the semantic space. The frame matching and occlusion handling are therefore delegated to be learned by the network.

Specifically, we design a transformer \cite{vaswani2017attention} style module to reason global correspondences between the current and previous frame features through a multihead attention layer followed by a 2D feed forward convolution. The attentive features are preserved for the inference of the next frame. Similar to other vision transformers, we encode positional information following the Cosine-Sine strategy, and add the 2D encodings to the current frame features to serve as query for the multihead attention. The preserved last frame features function as both value and key. The proposed semantic alignment module is placed at the deepest level of $\mathcal{F}_d$ for better efficiency. The design details are shown in Figure \ref{fig:operator}.

% Unlike other vision transformers that embed complex position information, we empirically found that the involvement of positional encodings does not help with the alignment process.

% learn an extra \textit{key vector} to assist the alignment.

\begin{figure*}[t]
	\begin{center}
		\includegraphics[width=1.0\linewidth]{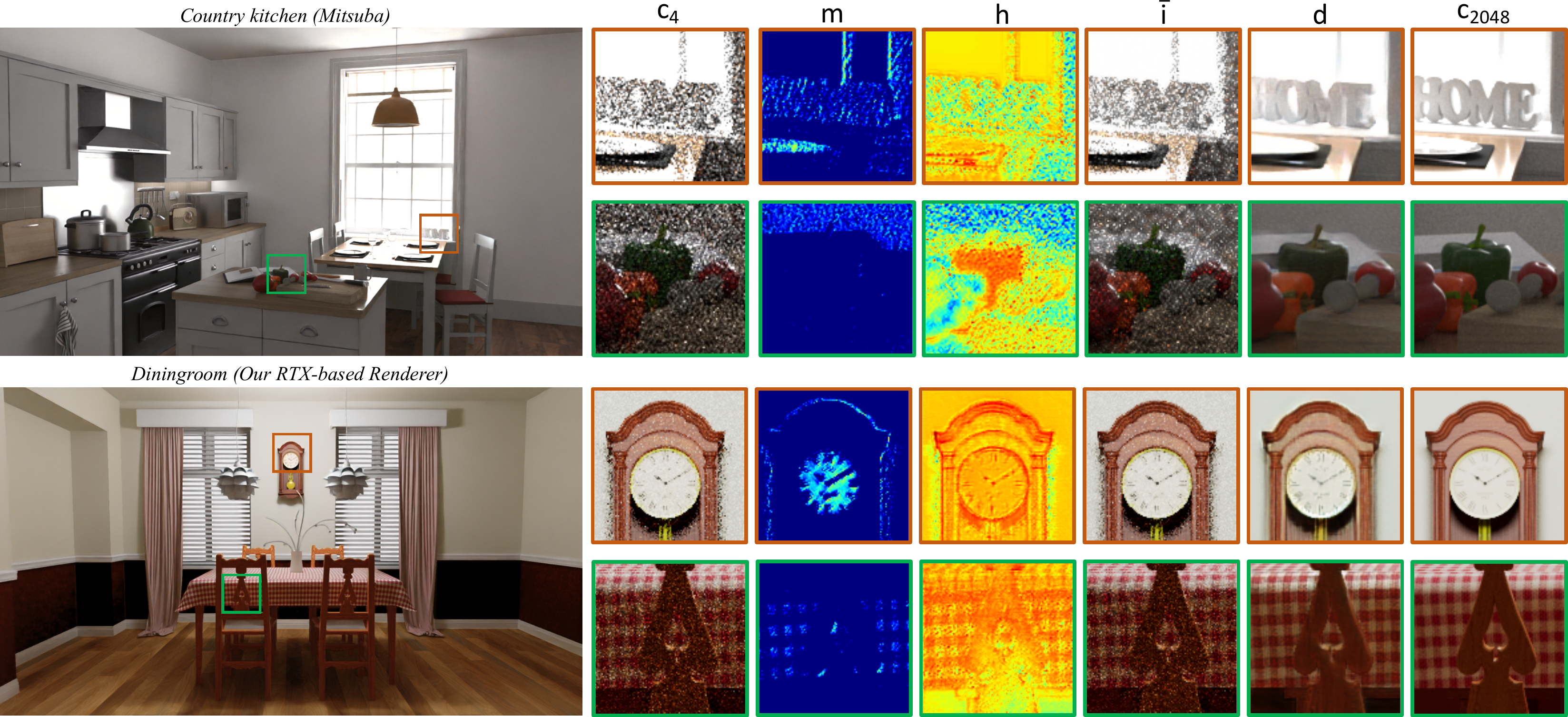}
	\end{center}
	\caption{\textbf{Qualitative results}. Left: full-scale denoising results. Right: zoomed in denoising details with intermediate results.}
	\label{fig:results}
\end{figure*}

% Unlike $\mathcal{L}_s$ that computes directly on the denoised frame $\mathbf{d}$ and the high spp reference $\mathbf{c}_{2048}$,
\noindent
\textbf{Loss functions.} Appropriate loss functions also help smooth the instabilities. We adopt simple $\textit{L}_1$ losses for both spatial and temporal constraints, denoted as $\mathcal{L}_s$ and $\mathcal{L}_t$ respectively. $\mathcal{L}_t$ is determined on $\Delta \mathbf{d}$ and $\Delta \mathbf{c}$ \cite{chaitanya2017interactive, hasselgren2020neural}, which can be pre-computed as $\Delta \mathbf{d}=\mathbf{d}^i-\mathbf{d}^{i-1}$ and $\Delta \mathbf{c}=\mathbf{c}^i_{2048}-\mathbf{c}^{i-1}_{2048}$ at current and previous frames.

Apart from the standard $\textit{L}_1$ constraints, we apply an additional perceptual loss term \cite{johnson2016perceptual} $\mathcal{L}_p$ to supervise networks based on the feature difference between $\mathbf{d}$ and $\mathbf{c}_{2048}$ inferred by a pre-trained VGG-16 network \cite{simonyan2014very}. Gradients are calculated on the weighted combination of the three losses:

\begin{equation}
    \mathcal{L}=\mathbf{W}_s \mathcal{L}_s + \mathbf{W}_t \mathcal{L}_t + \mathbf{W}_p \mathcal{L}_p,
\end{equation}
where $\mathbf{W}$ are scale factors applied on each loss. During training, $\mathcal{L}_s$ and $\mathcal{L}_p$ are only applied on the last frame in a video sequence, while $\mathcal{L}_t$ involves last two frames.

\section{Experiments}

\subsection{Experimental setup}

\noindent
\textbf{Datasets.} Our method was first benchmarked on the videos captured from 10 synthesized scenes \cite{resources16} (\textit{Contemporary Bathroom}, \textit{The Grey} \& \textit{White Room}, \textit{Bedroom}, \textit{Country Kitchen}, \textit{Modern Hall}, \textit{The Wooden Staircase}, \textit{Salle De Bain}, \textit{The White Room}, \textit{The Breakfast Room}, \textit{Japanese Classroom}) using the Mitsuba renderer \cite{Mitsuba}. The trained models were then slightly fine-tuned and transferred to validate on 3 new scenes (\textit{Warmroom}, \textit{Diningroom}, \textit{Livingroom}) rendered from our self-built RTX-based renderer. All scenes commonly have more than one light source and abundant environment illumination. We set up a camera to traverse through the scenes and captured videos based on a set of predefined key nodes in the scenes. Camera position and orientation were linear and quaternion interpolated between two adjacent key nodes by a specified frequency. As mentioned in Sec. \ref{preliminary}, we pre-rendered 5 color images at each camera position with spp in $2^{\{1,2,3,4,5\}}$ to simulate rendering of $[1,63]$ spp images. GBuffers (normal, depth, albedo) are obtained as the rasterization results at the same view. In total, 8 training images in $1920\times 1080$ resolution are required to be rendered at each camera location. 

\begin{table}[t]
		
		\centering
		\resizebox{\linewidth}{!}{\begin{tabular}{{c|c c| c c}}
			\toprule
			Methods & PSNR$\uparrow$ & SSIM$\uparrow$ & MACs & Params\\
			\hline
			\hline
			U-Net \cite{ronneberger2015u} & 31.91 & 86.00 & 316.52 & 0.972\\
			FastDVDnet \cite{Tassano_2020_CVPR} & 32.07& 86.02 & 1083.5 & 1.867\\
			RAE \cite{chaitanya2017interactive} & 31.96 & 86.07 & 432.61 & 1.864\\ 
			\hline
			DASR \cite{kuznetsov2018deep} & 31.97 & 86.08 & 558.00 & 1.577\\
			NTASAD \cite{hasselgren2020neural} & 32.13 & 86.46 & 826.11& 1.578\\
			\hline
			Ours & \textbf{33.39} & \textbf{87.65} & 478.22 & 1.285\\
			\bottomrule
		
		\end{tabular}}
		\caption{\textbf{Quantitative results on Mitsuba with 4 spp.} SSIM are in $\times10^{-2}$, MACs are in $\times10^9$, Params are in $\times10^6$. }\label{tab:mitsuba}
\end{table}

For Mitsuba scenes, we captured one video per scene with 500 frames using the provided camera trajectories in \cite{chaitanya2017interactive}. The first 100 frames and the last 400 frames are split for validation (testing) and training respectively. We further construct non-overlapping video clips with 5 consecutive frames to be fed in each network forward propagation during both training and validation, which results in 800 training clips and 200 validation clips. In testing, instead of using short clips, models are inferred on all 100 consecutive testing frames directly. For generalization experiments, videos were captured in 100 frames. 

\begin{figure}[t]
	\begin{center}
		\includegraphics[width=1.0\linewidth]{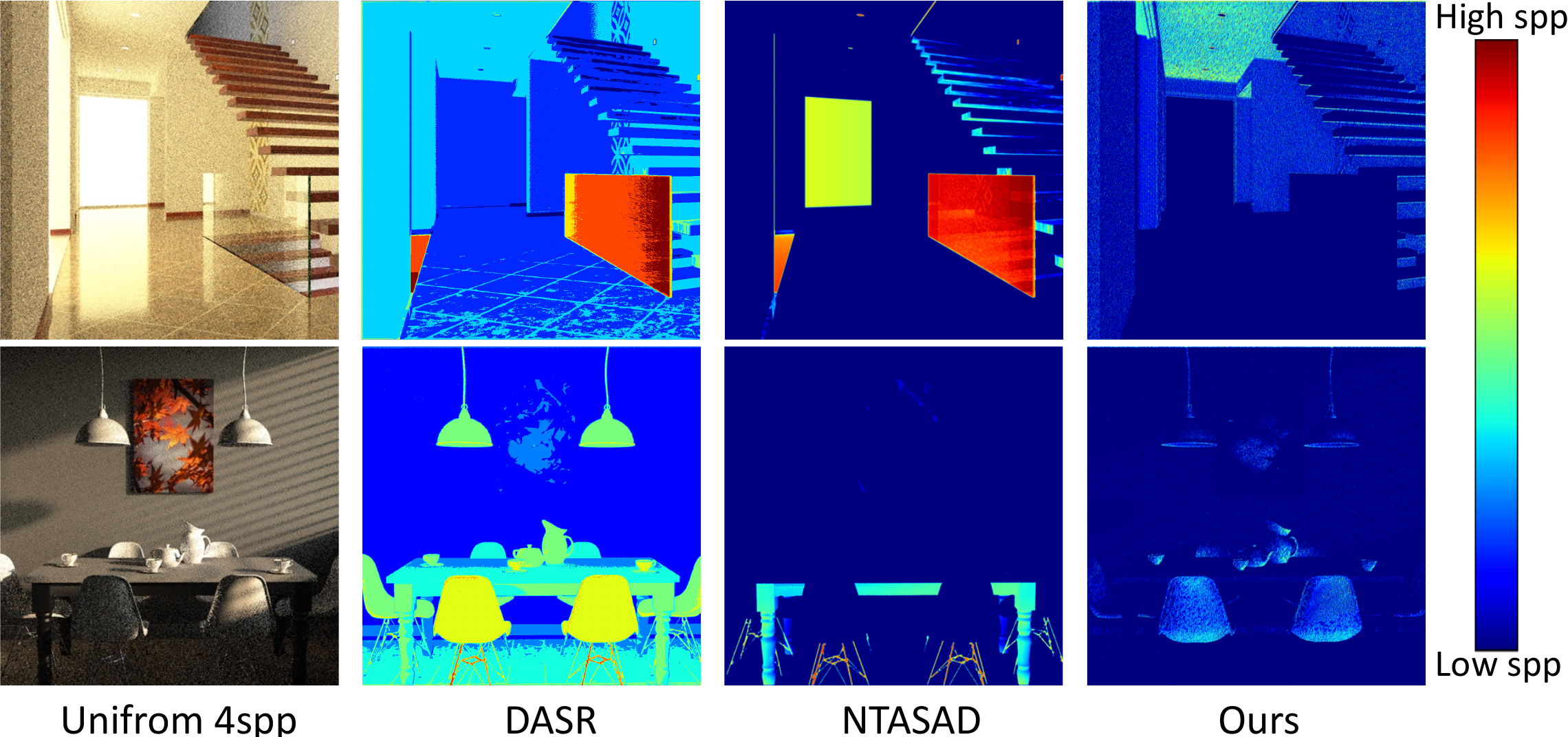}
	\end{center}
	\caption{\textbf{Comparisons of sampling maps}.}
	\label{fig:sample}
\end{figure}

\begin{figure*}[t]
	\begin{center}
		\includegraphics[width=1.0\linewidth]{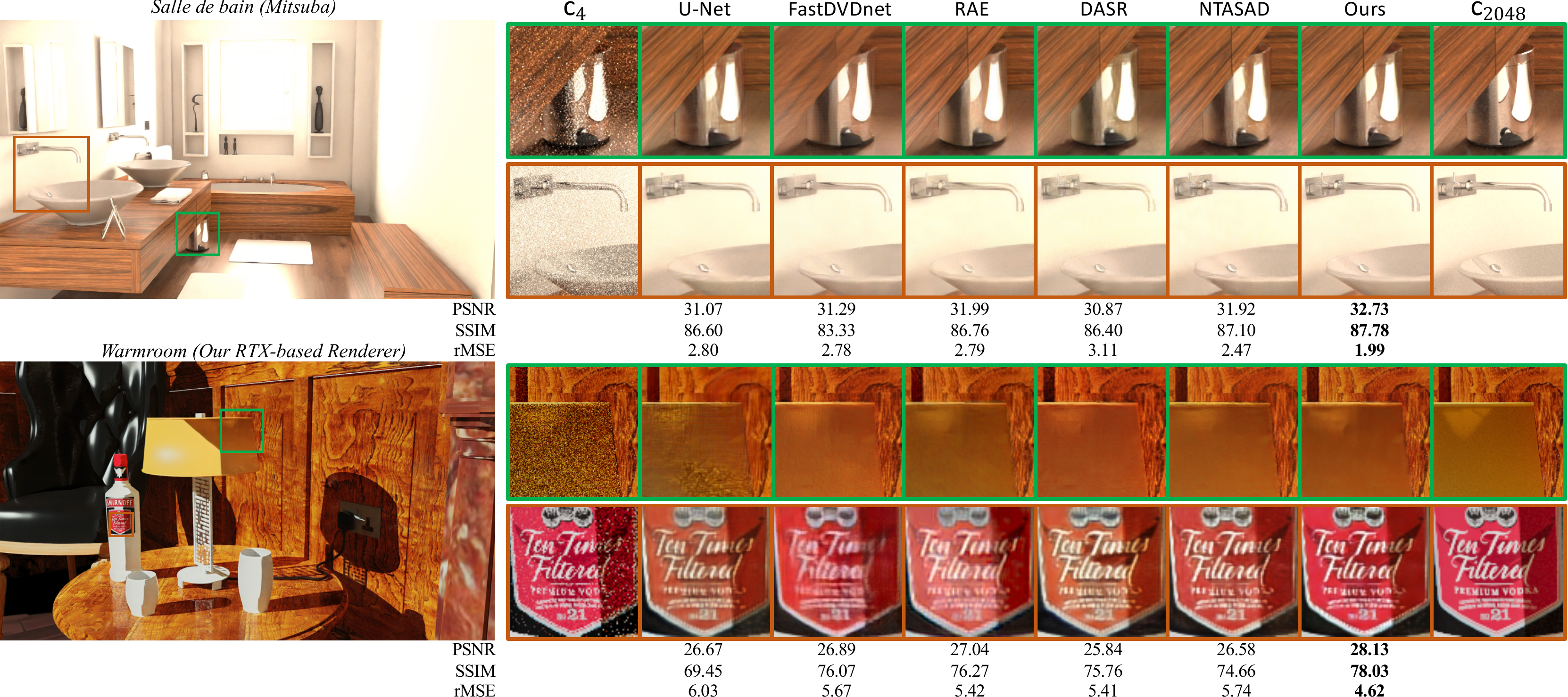}
	\end{center}
	\caption{\textbf{Qualitative comparisons}. Left: Our full-scale denoising results. Right: zoomed in denoising details of all competing methods.  Single frame metrics are reported as well. }
	\label{fig:compare}
\end{figure*}

\noindent
\textbf{Implementation details.} Adam optimizer \cite{kingma2014adam} was used to train all models starting from an initial learning rate of 3$e^{-4}$ and CosineAnealing scheduled to 3$e^{-6}$ in 300 epochs. We randomly cropped $256^2$ patches for training and center cropped $512^2$ patches for validation. After training converges, we retain the model instance at the epoch with the lowest validation loss to test on full-scale frames. Due to a large number of possible training patches, we didn't use extra data augmentation techniques. We set batch size to 8 for training and 1 for validation and testing. $\mathbf{W}_s$, $\mathbf{W}_t$ and $\mathbf{W}_p$ are set to 0.8, 0.2 and 0.1 respectively. Perceptual loss was also applied on all competing methods for fair comparison. The kernel pool with $q=128$ and $l=5$ was binarily initialized following Bernoulli distribution. All experiments were implemented in Pytorch \cite{paszke2019pytorch} and on Tesla V100 GPUs.

Unless explicitly specified, we expect an average spp budget of 4 for inputs and uniform 2048 spp for reference. Following \cite{hasselgren2020neural}, color images (albedo and $\mathbf{c}_1$) are converted into gray scale as two additional inputs for $\mathcal{F}_s$. During network forward, $\mathbf{c}_1$ and $\mathbf{i}$ requires two independent renderings. For consistent rendering costs, we subtract 1 spp from the expected budget and linearly blend $\mathbf{c}_1$ to $\mathbf{i}$.

\subsection{Results}

\noindent
\textbf{Mitsuba results.} We first present our denoising results on the videos captured in Mitsuba path traced scenes. Figure \ref{fig:results} shows qualitative results on one exemplary frame. In $\mathbf{h}$, we observe that the interpolation principle of kernels is based not only on pixel chrom, but also the semantics reasoned in $\mathcal{F}_s$. Regions with the same color can be denoised with very different kernels. As shown in $\mathbf{m}$, our framework demonstrates a good ability to collaborate both adaptive sampling and initial denoising by preserving more samples to the hard-to-denoise regions and fewer samples to the easier ones. We compare our sampling map to other adaptive sampling methods' in Figure \ref{fig:sample}. Interestingly, our sampling maps significantly differ from \cite{kuznetsov2018deep, hasselgren2020neural} which always assign high spp to specific objects (e.g. glass, table). Our adaptive sampling and initial denoising procedure serve as good complements of each other and the produced $\bar{\mathbf{i}}$ becomes more comprehensible to $\mathcal{F}_d$. The final denoised images prove that our method is not only able to remove the realistic noise, but is also capable of reconstructing details (e.g. highlight, object geometry).

We further compare our method to the baseline method: U-Net \cite{ronneberger2015u}; state-of-the-art videos denoisers: FastDVDnet \cite{Tassano_2020_CVPR}, RAE \cite{chaitanya2017interactive}; and adaptive sampling counterparts: DASR \cite{kuznetsov2018deep}, NTASAD \cite{hasselgren2020neural}. The average quantitative results on the entire 1000 frame testing set are shown in Table \ref{tab:mitsuba}. Our proposed framework outperforms all competing methods on both metrics greatly. Notably, except for the methods that use simple encoder-decoder structure \cite{ronneberger2015u, chaitanya2017interactive}, our dual-network framework demands the least computation costs and parameter counts. Exemplary qualitative comparisons are shown in Figure \ref{fig:compare} Top, and more cases are shown in the supplementary material.

\begin{figure*}
\centering
\begin{minipage}[]{0.8\textwidth}
\centering
    \resizebox{1.0\linewidth}{!}{\begin{tabular}{{c|c c c c c c | c c c c | c c}}
			\toprule
			Model & UN & AS & KP & PP & FG & SA & PSNR$\uparrow$ & SSIM$\uparrow$ & $\mathcal{L}_t\downarrow$ & rMSE$\downarrow$& MACs & Params\\
			\hline
			\hline
			A & \checkmark &&&&& & 31.92 & 86.00 & 2.51 & 5.58 & 316.52 & 0.972\\
			B & \checkmark & \checkmark &&&& & 32.13 & 86.81 & 2.52 & 5.09 & 464.87 & 1.348\\
			C & \checkmark & \checkmark & \checkmark &&&& 33.11 & 87.31 & 2.27 & 4.03 & 526.05 & 1.471\\
			D & \checkmark & \checkmark & \checkmark & \checkmark &&& 33.29 & 87.51 & 2.17 & 3.79 & 526.40 & 1.471\\
			E & \checkmark & \checkmark & \checkmark & \checkmark & \checkmark && 33.34 & 87.56 & 2.18 & 3.70 & 478.54 & 1.324\\
			F & \checkmark & \checkmark & \checkmark & \checkmark & \checkmark & \checkmark & \textbf{33.39} & \textbf{87.65} & \textbf{2.16} & \textbf{3.67} & 478.22 & 1.285\\
			\hline
			G & \checkmark & \checkmark & \checkmark & & \checkmark && 33.22 & 87.33 & 2.20 & 3.89 & 478.34 & 1.324\\
			H & \checkmark & \checkmark & \checkmark & & \checkmark & \checkmark & 33.19 & 87.36 & 2.22 & 3.90 & 478.02 & 1.284\\
			I & \checkmark & \checkmark & \checkmark & \checkmark & & \checkmark & 33.15 & 87.40 & 2.22 & 4.00 & 525.94 & 1.432\\
			J & \checkmark & \checkmark & \checkmark & & & \checkmark & 33.14 & 87.29 & 2.25 & 4.06 & 525.74 & 1.432\\
			\hline
			K & \checkmark & \checkmark & & \checkmark & & & 32.46 & 86.94 & 2.43 & 4.73 & 465.07 & 1.348\\
			L & \checkmark & \checkmark & & & \checkmark & & 32.43 & 86.88 & 2.43 & 4.72 & 439.43 & 1.269\\
			M & \checkmark & \checkmark & & & & \checkmark & 32.39 & 87.02 & 2.48 & 4.68 & 464.56 & 1.309\\
			\bottomrule
		\end{tabular}}
		%\captionof{table}{yes}
\end{minipage}%
\hfill
\begin{minipage}[c]{0.18\textwidth}
\raggedleft
	\includegraphics[width=1.\textwidth]{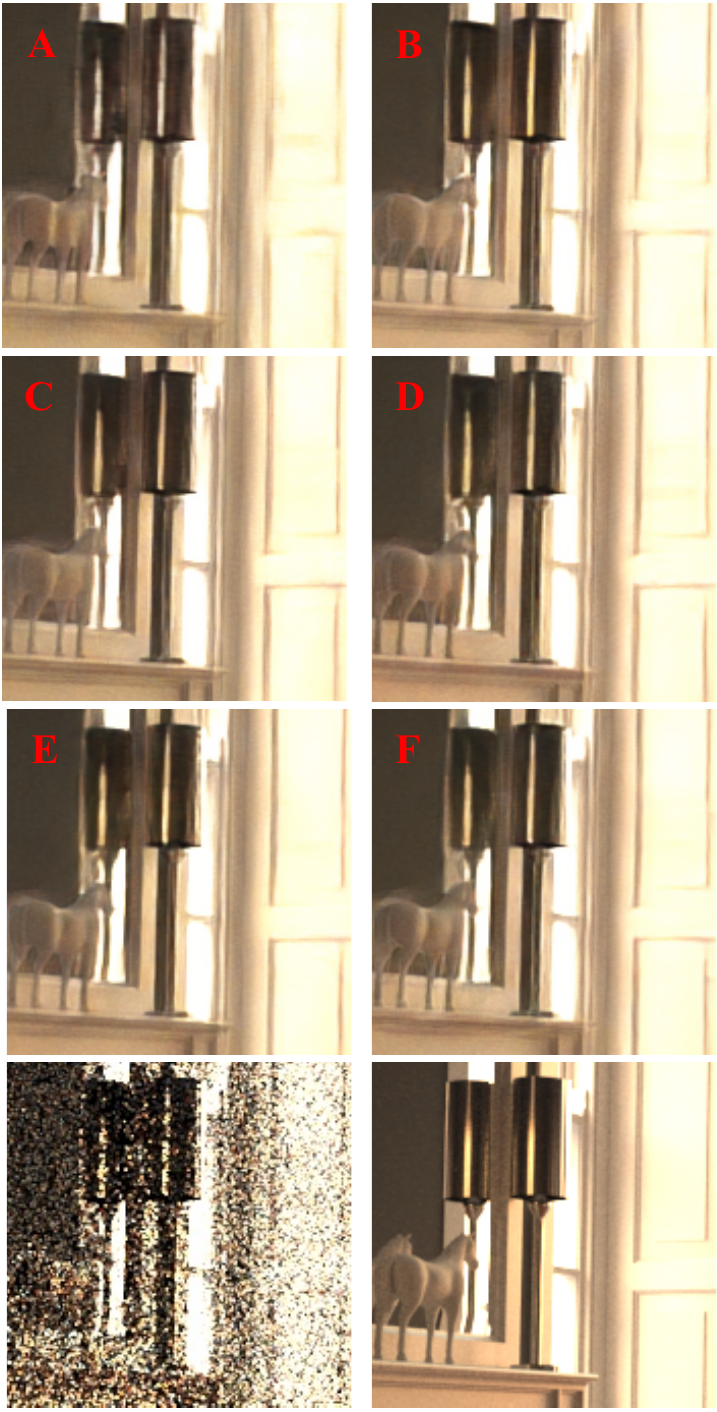}
	%\captionof{figure}{hello}
\end{minipage}
\caption{\textbf{Component study}. \textbf{Left:} UN denotes using $\mathcal{F}_d$ only, AS triggering Adaptive Sampling strategy with $\mathcal{F}_s$, KP promoting to two-stage denoising with Kernel Pool, PP the usage of Position-aware Pooling, FG the usage of Fast GhostConv, SA the usage of Semantic Alignment. SSIM are in $\times10^{-2}$, rMSE are in $\times10^{-3}$, $\mathcal{L}_t$ are in $\times10^{-2}$, MACs are in $\times10^9$, Params are in $\times10^6$. \textbf{Right:} Qualitative comparisons of model A-F, uniform 4spp image and reference are shown at the bottom.} \label{component}
\end{figure*}

% The sampling map $\mathbf{m}$ and kernel map $\mathbf{h}$ generated by our $\mathcal{F}_s$ reveal different values for different regions

\begin{figure*}[t]
	\begin{center}
		\includegraphics[width=1.0\linewidth]{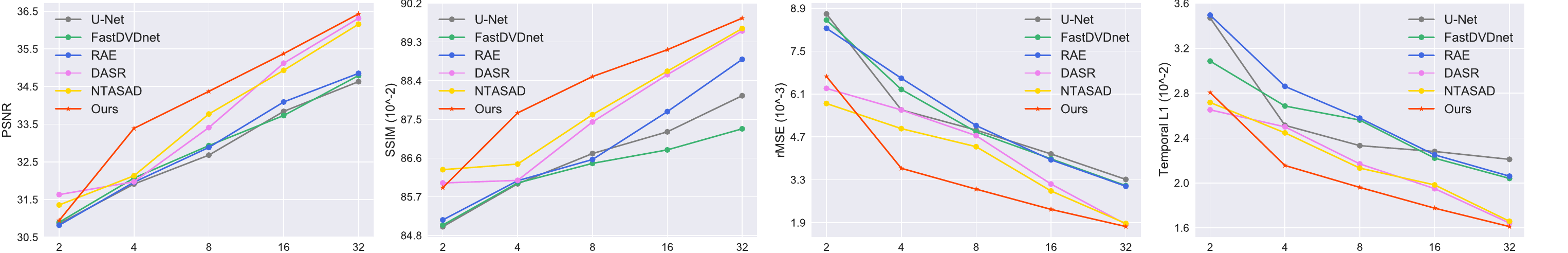}
	\end{center}
	\caption{\textbf{Denoising on different spp images.} Metrics including: PSNR, SSIM, rMSE and $\mathcal{L}_t$ are reported.}
	\label{fig:spp}
\end{figure*}

% In Figure \ref{fig:sample}, we visualize the sampling maps of the adaptive sampling methods. Interestingly, we found that our sampling map is not sensitive to object attributes, differing from \cite{kuznetsov2018deep, hasselgren2020neural} that always assign high spp to specific objects (e.g. glass, table). Our sampling map complements the initial denoising process by assigning more samples to the poorly denoised regions and less samples to the good ones, therefore, the construction of $\bar{\mathbf{i}}$ achieves a good balance on utilizing both adaptive sampling and kernel pool denoising. 

\noindent
\textbf{Generalization to the new renderer.} Monte Carlo denoisers are normally trained in a renderer-dependent basis and generalizing to new scenes rendered using another renderer can be challenging. Here we conducted extensive experiments by validating the trained denoisers on three scenes rendered from our self-built RTX-based renderer. Since the two renderers use distinct color palettes and GBuffers are rasterized in different scales, a direct application of the existing models to the new scenes is infeasible. To solve the inconsistency, we first manually align the scales of the GBuffers that are then used to fine-tune the trained models slightly for 30 epochs on each of the three new scenes using a single 5-frame clip only. The fine-tuning frames are captured separately apart from the 100 testing frames. All competing methods were fine-tuned followed the same procedure for fair comparisons. Our denoising results are shown in Figure \ref{fig:results} bottom and compared in Figure \ref{fig:compare} Bottom. Although all methods demonstrate limited generalization ability, our method stands out with superior reconstruction performances on texture and lighting details.

% Surprisingly, we found that the adaptive sampling methods \cite{kuznetsov2018deep, hasselgren2020neural} yield limited generalization ability after the mild fine-tuning, and the denoising qualities are poorer than the ones inferred by standard video denoisers \cite{ronneberger2015u, Tassano_2020_CVPR, chaitanya2017interactive}. However, our initial stage denoising works as a good remedy and achieves superior denoising qualities.

\subsection{Ablation study}

In this subsection, we first analyze the effectiveness of each of the framework components and then validate the methods on different rendering circumstances with different spp budgets. More extensive ablation studies are presented in the supplementary material.

\noindent
\textbf{Component study.} The impact of the individual component of our framework was examined by simply removing it from the full framework. We report two more metrics: temporal $L_1$ loss and rMSE \cite{rousselle2011adaptive} for a more comprehensive analysis in Figure \ref{component} left. Model A-F list the experiments of adding one module to the existing framework at a time. Model C shows that the proposed two-stage denoiser promotion with kernel pool yields the greatest performance gain. When equipping all proposed modules (model F), our method achieves the best performance in terms of all metrics. Model G-J demonstrate the impact of each instability smoothing module building on the two-stage denoising basis. All modules can bring further performance improvement. Model complexity can be even reduced when FG and SA are enabled (model H). Model K-M validate the capability of the proposed modules when disabling two-stage denoising. The usage of PP alleviates spatial flickering which impacts the visual quality directly and brings the greatest improvements on all metrics (model K). Qualitative results of the component study are shown in Figure \ref{component} right. 

\noindent
\textbf{Denoising on different spp images.} We study the feasibility of applying denoising methods on noisy images rendered with a wide range of spps ($2^{\{1,2,3,4,5\}}$). Average spp budget was set for adaptive sampling methods, while uniformly sampled images rendered with the same spp were fed to non-adaptive sampling methods. The results are presented in Figure \ref{fig:spp}. Undoubtedly, our adaptive sampling based framework performs considerably better than standard video denoisers on denoising Monte Carlo path traced images regardless of spp settings. However, When denoising images with an extremely low spp budget, we found our two-stage denoising framework hardly shows any advantages compared to adaptive sampling counterparts. When spp budget is greater than 2, our method demonstrates superior performances that surpass all competing methods on all metrics.

\section{Conclusion}

In this paper, we proposed a novel two-stage denoising framework for Monte Carlo path tracer with adaptive sampling. Concurrent to generating sampling maps, our framework produces additional kernel maps to interpolate per pixel kernels in a public kernel pool achieving an initial denoising. The resulting image is then fed into the denoising network with the proposed position-aware pooling for alleviating spatial flickering. We then design the semantic alignment technique to handle temporal instabilities by aligning adjacent frames in the semantic space. Our method was validated on noisy videos captured in Mitsuba path traced scenes and the scenes rendered in our RTX-based path tracer. Under the same experimental settings, our method surpasses state-of-the-art counterparts qualitatively and quantitatively with the least computational costs.

{\small
\bibliographystyle{ieee_fullname}
\bibliography{egbib}
}

\end{document}